\documentclass{article}
\usepackage{ijcai21}

\usepackage{times}
\usepackage{soul}
\usepackage{url}
\usepackage[hidelinks]{hyperref}
\usepackage[utf8]{inputenc}
\usepackage[small]{caption}
\usepackage{graphicx}
\usepackage{amsmath}
\usepackage{amsthm}
\usepackage{booktabs}
\usepackage{algorithm}
\usepackage{algorithmic}
\usepackage{multirow}
\usepackage{amsmath}
\usepackage{amsfonts}
\usepackage{booktabs}
\usepackage{appendix}

\newcommand{\citep}{\cite}
\newcommand{\citet}[1]{\citeauthor{#1}~\shortcite{#1}}

\title{Coarse-to-Fine Entity Representations for Document-level Relation Extraction}

\author{
    Damai Dai\textsuperscript{1}\thanks{Equal Contribution.}
    \and
    Jing Ren\textsuperscript{1}\footnotemark[1]
    \and
    Shuang Zeng\textsuperscript{1}
    \and
    Baobao Chang\textsuperscript{1,2}
    \and
    Zhifang Sui\textsuperscript{1,2}
    \affiliations
    \textsuperscript{1}MOE Key Lab of Computational Linguistics, School of EECS, Peking University \\
    \textsuperscript{2}Peng Cheng Laboratory, China
    \emails
    \{daidamai,rjj,zengs,chbb,szf\}@pku.edu.cn
}

\begin{document}

\maketitle

\begin{abstract}

Document-level Relation Extraction (RE) requires extracting relations expressed within and across sentences. 
Recent works show that graph-based methods, usually constructing a document-level graph that captures document-aware interactions, can obtain useful entity representations thus helping tackle document-level RE. 
These methods either focus more on the entire graph, or pay more attention to a part of the graph, e.g., paths between the target entity pair. 
However, we find that document-level RE may benefit from focusing on both of them simultaneously. 
Therefore, to obtain more comprehensive entity representations, we propose the \textbf{C}oarse-to-\textbf{F}ine \textbf{E}ntity \textbf{R}epresentation model (\textbf{CFER}) that adopts a coarse-to-fine strategy involving two phases. 
First, CFER uses graph neural networks to integrate global information in the entire graph at a coarse level. 
Next, CFER utilizes the global information as a guidance to selectively aggregate path information between the target entity pair at a fine level. 
In classification, we combine the entity representations from both two levels into more comprehensive representations for relation extraction. 
Experimental results on two document-level RE datasets, DocRED and CDR, show that CFER outperforms existing models and is robust to the uneven label distribution.\footnote{Our code and data are available at \url{https://github.com/Hunter-DDM/cfer-document-level-RE}. }

\end{abstract}

\section{Introduction}

Relation Extraction (RE) aims to extract semantic relations between named entities from plain text. 
It is an efficient way to acquire structured knowledge automatically, thus benefiting various natural language processing (NLP) applications, especially knowledge graph construction~\citep{scierc}. 
Most of the previous RE works focus on the sentence level, i.e., they extract the relations within only a single sentence~\citep{cnn_re,bilstm_re}. 
However, in real-world scenarios, sentence-level RE models may omit some inter-sentence relations while a considerable number of relations are expressed beyond a single sentence but across multiple sentences in a long document~\citep{docred}. 
Therefore, document-level RE has attracted much attention in recent years. 

\begin{figure}[t]
\centering
\includegraphics[width=0.99\linewidth]{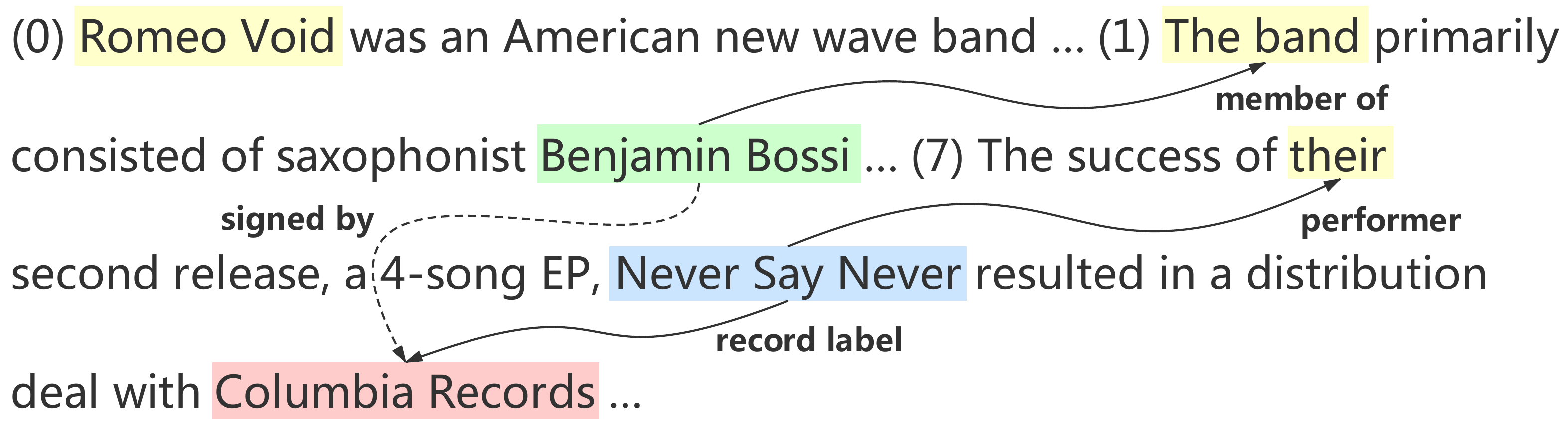}
\caption{
An example for document-level RE. 
Word spans with background colors denote the mentions. 
Mentions that refer to the same entity have the same background color.
The solid lines denote intra-sentence relations. 
The dotted line denotes an inter-sentence relation. 
Document-level RE requires extracting both intra- and inter-sentence relations between all the target entity pairs. 
}
\label{fig:task}
\end{figure}

Figure~\ref{fig:task} shows an example document and corresponding relational facts for document-level RE. 
In the example, to extract the relation between \textit{Benjamin Bossi} and \textit{Columbia Records}, two entities separated by several sentences, we need the following inference steps. 
First, we need to know that \textit{Benjamin Bossi} is a member of \textit{Romeo Void}. 
Next, we need to infer that \textit{Never Say Never} is performed by \textit{Romeo Void}, and released by \textit{Columbia Records}. 
Based on these facts, we can draw a conclusion that \textit{Benjamin Bossi} is signed by \textit{Columbia Records}. 
The example indicates that, to tackle document-level RE, a model needs the ability to capture interactions between long-distance entities. 
In addition, since a document may have an extremely long text, a model also needs the ability to integrate global contextual information for words. 

\begin{figure*}[t]
\centering
\includegraphics[width=0.99\linewidth]{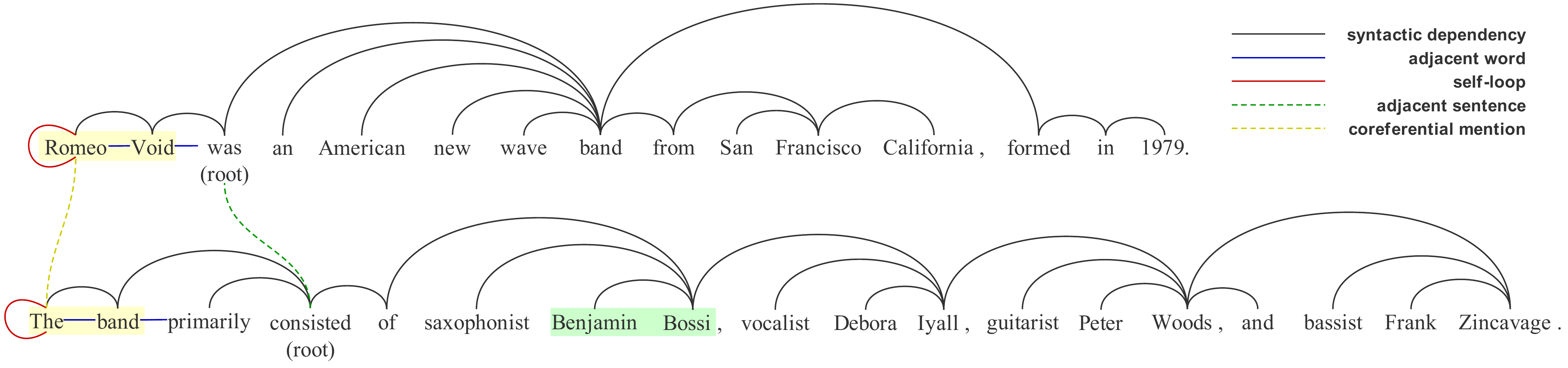}
\caption{
An illustration of a document-level graph corresponding to a two-sentence document. 
Each node in the graph corresponds to a word in the document. 
We design five categories of edges to connect nodes in the graph. 
For the simplicity of the illustration, we omit some self-loop edges and adjacent word edges. 
}
\label{fig:graph}
\end{figure*}
 
Recent works show that graph-based methods can obtain useful entity representations thus helping tackle document-level RE. 
These methods usually construct a document-level graph, which represents words as nodes and uses edges between them to capture document-aware interactions. 
To obtain entity representations for relation extraction, some of them use  graph neural networks (GNN)~\citep{labelled_edge_gcn,dcgcn} to integrate neighborhood information for each node~\citep{gcnn,lsr}. 
Although they consider the entire graph structure, they may fail to model the interactions between long-distance entities due to the inherent over-smoothing problem in GNN~\citep{over_smoothing}. 
Other works attempt to encode path information between the target entity pair in the graph~\citep{quirk_and_poon,eog}. 
They have the ability to alleviate the problem of modeling long-distance entity interactions, but they may fail to capture global contextual information since they usually integrate only local contextual information for nodes in the graph. 
Therefore, to obtain more comprehensive entity representations, it is necessary to find a way to integrate global contextual information and model long-distance entity interactions simultaneously. 

In this paper, we propose the \textbf{C}oarse-to-\textbf{F}ine \textbf{E}ntity \textbf{R}epresentation model (\textbf{CFER}) to obtain comprehensive entity representations for document-level RE. 
More specifically, we first construct a document-level graph that captures rich document-aware interactions, reflected by syntactic dependencies, adjacent word connections, cross-sentence connections, and coreferential mention interactions. 
Based on the constructed graph, we design a coarse-to-fine strategy with two phases. 
First, we use Densely Connected Graph Convolutional Networks (DCGCN)~\citep{dcgcn} to integrate global contextual information in the entire graph at a coarse level. 
Next, we adopt an attention-based path encoding mechanism, which takes the global contextual information as a guidance, to selectively aggregate path information between the potentially long-distance target entity pair at a fine level. 
Given the entity representations from both two levels that feature global contextual information and long-distance interactions, respectively, we can obtain more comprehensive entity representations for relation extraction by combining them. 

Our contributions are summarized as follows: 
\begin{itemize}
    \item 
    We propose a novel document-level RE model called \textbf{CFER}. 
    It uses a coarse-to-fine strategy to integrate global contextual information and model long-distance interactions between the target entities simultaneously, thus obtaining comprehensive entity representations. 
    
    \item 
    Experimental results on two popular document-level RE datasets, DocRED and CDR, show that CFER achieves better performance than existing models. 
    
    \item 
    Elaborate analysis validates the effectiveness of our coarse-to-fine strategy. 
    Further, we highlight the robustness of CFER to the uneven label distribution and the ability of CFER to model long-distance interactions. 
\end{itemize}

\section{Methodology}

\subsection{Task Formulation}

Let $\mathcal{D}$ denote a document consisting of $N$ sentences $\mathcal{D}=\left\{s_i\right\}^{N}_{i=1}$, 
where $s_i=\left\{w_j\right\}^{M}_{j=1}$ denotes the $i$-th sentence containing $M$ words denoted by $w_j$. 
Let $\mathcal{E}$ denote an entity set containing $P$ entities $\mathcal{E}=\left\{e_i\right\}^{P}_{i=1}$, where $e_i=\left\{m_j\right\}^{Q}_{j=1}$ denotes the coreferential mention set of the $i$-th entity, containing $Q$ word spans of corresponding mentions denoted by $m_j$. 
Given $\mathcal{D}$ and $\mathcal{E}$, document-level RE requires extracting all the relational facts in the form of triplets, i.e., extracting $\left\{(e_i, r, e_j) | e_i, e_j \in \mathcal{E}, r \in \mathcal{R} \right\}$, where $\mathcal{R}$ is a pre-defined relation category set. 
Since an entity pair may have multiple semantic relations, we formulate document-level RE as a multi-label classification task. 

\begin{figure*}[t]
\centering
\includegraphics[width=0.99\linewidth]{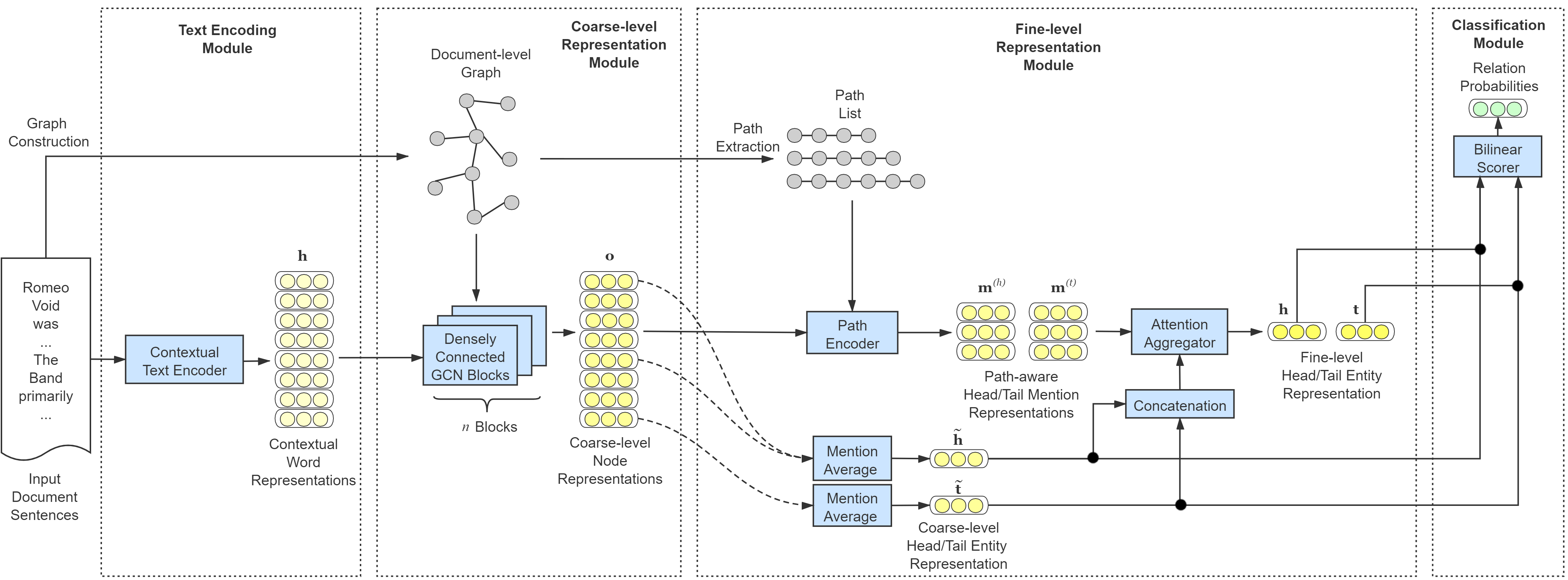}
\caption{
An illustration of our proposed model, CFER. 
It is composed of four modules: a text encoding module, a coarse-level representation module, a fine-level representation module, and a classification module. 
}
\label{fig:model}
\end{figure*}

\subsection{Document-level Graph Construction}
Given a document, we first construct a document-level graph that captures rich document-aware interactions, reflected by syntactic dependencies, adjacent word connections, cross-sentence connections, and coreferential mention interactions. 
Figure~\ref{fig:graph} shows an example document-level graph corresponding to a two-sentence document. 
The graph regards words in the document as nodes and captures document-aware interactions by five categories of edges. 
These undirected edges are described as follows. 

\noindent\textbf{Syntactic dependency edge:}
Syntactic dependency information is proved effective for document-level or cross-sentence RE in previous works~\citep{gcnn}. 
Therefore, we use the dependency parser in spaCy\footnote{https://spacy.io/} to parse the syntactic dependency tree for each sentence. 
After that, we add edges between all node pairs that have dependency relations. 

\noindent\textbf{Adjacent word edge:}
\citet{quirk_and_poon} point out that adding edges between adjacent words can mitigate the dependency parser errors.
Therefore, we add edges between all node pairs that are adjacent in the document. 

\noindent\textbf{Self-loop edge:}
For a node, in addition to its neighborhood information, the historical information of the node itself is also essential in information integration. 
Therefore, we add a self-loop edge for each node. 

\noindent\textbf{Adjacent sentence edge:}
To ensure that information can be integrated across sentences, for each adjacent sentence pair, we add an edge between their dependency tree roots. 

\noindent\textbf{Coreferential mention edge:}
Coreferential mentions may share information captured from their respective contexts with each other. 
This could be regarded as global cross-sentence interactions. 
Therefore, we add edges between the first words of all mention pairs that refer to the same entity. 

\subsection{Coarse-to-Fine Entity Representations}

In this subsection, we describe our proposed \textbf{C}oarse-to-\textbf{F}ine \textbf{E}ntity \textbf{R}epresentation model (\textbf{CFER}) in detail. 
As shown in Figure~\ref{fig:model}, our model is composed of a text encoding module, a coarse-level representation module, a fine-level representation module, and a classification module.  

\noindent\textbf{Text Encoding Module:} 
This module aims to encode each word in the document as a vector with text contextual information. 
By default, CFER uses GloVe~\citep{glove} embeddings and a Bi-GRU model~\citep{gru,bi_rnn} as the encoder. 
To improve the ability of contextual text encoding, we can also replace this module by Pre-Trained Models (PTM) such as BERT~\citep{bert} or RoBERTa~\citep{roberta}. 
This module finally outputs contextual word representations $\mathbf{h}_{i}$ of each word in the document: 
\begin{equation}
    \mathbf{h}_{i} = \text{Encoding} (\mathcal{D}),
\end{equation}
where $\mathbf{h}_{i} \in \mathbb{R}^{d_h}$, and $d_h$ denotes the hidden dimension. 

\noindent\textbf{Coarse-level Representation Module:} 
This module aims to integrate local and global contextual information in the entire document-level graph. 
As indicated by \citet{dcgcn}, Densely Connected Graph Convolutional Networks (DCGCN) have the ability to capture rich local and global contextual information. 
Therefore, we adopt DCGCN layers as the coarse-level representation module. 
DCGCN layers are organized into $n$ blocks and the $k$-th block has $m_k$ sub-layers. 
At the $l$-th sub-layer in block $k$, the calculation for node $i$ is defined as 
\begin{align}
    \mathbf{h}_i^{(k, l)} &= \text{ReLU} \left( \sum_{j \in \mathcal{N}(i)} \mathbf{W}^{(k, l)} \hat{\mathbf{h}}_{j}^{(k, l)} + \mathbf{b}^{(k, l)} \right), \\
    \hat{\mathbf{h}}_{j}^{(k, l)} &= [\mathbf{x}_j^{(k)}; \mathbf{h}_j^{(k, 1)}; ...; \mathbf{h}_j^{(k, l-1)}], 
\end{align}
where $\mathbf{x}_j^{(k)} \in \mathbb{R}^{d_h}$ is the block input of node $j$. 
$\mathbf{h}_i^{(k, l)} \in \mathbb{R}^{d_h/m_k}$ is the output of node $i$ at the $l$-th sub-layer in block $k$. 
$\hat{\mathbf{h}}_{j}^{(k, l)} \in \mathbb{R}^{d_h + (l - 1) d_h/m_k}$ is the neighborhood input, obtained by concatenating $\mathbf{x}_j^{(k)}$ and the outputs of node $j$ from all previous sub-layers in the same block. 
$\mathbf{W}^{(k, l)}$ and $\mathbf{b}^{(k, l)}$ are trainable parameters. 
$\mathcal{N}(i)$ denotes the neighbor set of node i in the document-level graph. 
Finally, block $k$ adds up the block input and the concatenation of all sub-layer outputs, and then adopts a fully connected layer to compute the block output $\mathbf{o}_i^{(k)} \in \mathbb{R}^{d_h}$: 
\begin{equation}
    \mathbf{o}_i^{(k)} = \text{FC} \left(\mathbf{x}_i^{(k)} + [\mathbf{h}_i^{(k, 1)}; ...; \mathbf{h}_i^{(k, m_k)}]\right). 
\end{equation}
Finally, we take the output of the final block $\mathbf{o}_i^{(n)}$ as the output of the coarse-level representation module. 

\noindent\textbf{Fine-level Representation Module:} 
The coarse-level representations can capture rich contextual information, but may fail to model long-distance entity interactions. 
Taking the global contextual information as a guidance, the fine-level representation module aims to utilize path information between the target entity pair to alleviate this problem. 
This module adopts an attention-based path encoding mechanism based on a path encoder and an attention aggregator. 

For a target entity pair $(e_1, e_2)$, we denote the numbers of their corresponding mentions as $\left|e_1\right|$ and $\left|e_2\right|$, respectively. 
We first extract $\left|e_1\right| \times \left|e_2\right|$ shortest paths between all mention pairs in a subgraph that contains only syntactic dependency and adjacent sentence edges. 
For the $i$-th path $[w_1, ..., w_{{len}_i}]$,  we then adopt a Bi-GRU model as the path encoder to compute the path-aware mention representations: 
\begin{align}
    \overrightarrow{\mathbf{p}_{i, j}} &= \overrightarrow{\mathbf{GRU}} \left( \overrightarrow{\mathbf{p}_{i, j-1}}, \mathbf{o}^{(n)}_{w_j} \right), \\
    \overleftarrow{\mathbf{p}_{i, j}} &= \overleftarrow{\mathbf{GRU}} \left( \overleftarrow{\mathbf{p}_{i, j+1}}, \mathbf{o}^{(n)}_{w_j} \right), \\
    \mathbf{m}^{(h)}_{i} &= \overleftarrow{\mathbf{p}_{i, 1}}, \quad
    \mathbf{m}^{(t)}_{i} = \overrightarrow{\mathbf{p}_{i, {len}_i}}, 
\end{align}
where $\overrightarrow{\mathbf{p}_{i, j}}, \overleftarrow{\mathbf{p}_{i, j}} \in \mathbb{R}^{d_h}$ are the forward and backward GRU hidden states of the $j$-th node in the $i$-th path, respectively. 
$\mathbf{m}^{(h)}_{i}, \mathbf{m}^{(t)}_{i} \in \mathbb{R}^{d_h}$ are the path-aware representations of the head and tail mentions in the $i$-th path, respectively. 

Since not all the paths contain useful information, we design an attention aggregator, which takes the global contextual information as a guidance, to selectively aggregate the path-aware mention representations: 
\begin{align}
    \mathbf{\widetilde{h}} &= \frac{1}{\left|e_1\right|} \sum_{j \in e_1} {\mathbf{o}^{(n)}_{j}}, \quad
    \mathbf{\widetilde{t}} = \frac{1}{\left|e_2\right|} \sum_{j \in e_2} {\mathbf{o}^{(n)}_{j}},  \\
    \alpha_i &= \mathop{\text{Softmax}}\limits_{i} \left( \mathbf{W}_a [\mathbf{\widetilde{h}}; \mathbf{\widetilde{t}}; \mathbf{m}^{(h)}_{i}; \mathbf{m}^{(t)}_{i}] + \mathbf{b}_a \right), \\
    \mathbf{h} &= \sum_i {\mathbf{m}^{(h)}_{i} \cdot \alpha_i}, \quad
    \mathbf{t} = \sum_i {\mathbf{m}^{(t)}_{i} \cdot \alpha_i}, 
\end{align}
where $\mathbf{\widetilde{h}},\mathbf{\widetilde{t}} \in \mathbb{R}^{d_h}$ are the coarse-level head and tail entity representations, respectively, which are computed by averaging their corresponding coarse-level mention representations. 
$\mathbf{W}_a$ and $\mathbf{b}_a$ are trainable parameters. 
$\mathbf{h}, \mathbf{t} \in \mathbb{R}^{d_h}$ are the path-aware fine-level representations of the head and tail entities, respectively. 

\noindent\textbf{Classification Module:}
In this module, we combine the entity representations from both two levels to obtain comprehensive representations that capture both global contextual information and long-distance entity interactions. 
Next, we predict the probability of each relation by a bilinear scorer: 
\begin{equation}
    P(r|e_1, e_2) = \text{Sigmoid} \left( [\mathbf{\widetilde{h}}; \mathbf{h}]^{\text{T}} \mathbf{W}_c [\mathbf{\widetilde{t}}; \mathbf{t}] + \mathbf{b}_c \right)_r, 
\end{equation}
where $\mathbf{W}_c \in \mathbb{R}^{2d_h \times n_r \times 2d_h} $ and  $\mathbf{b}_c \in \mathbb{R}^{n_r}$ are trainable parameters with $n_r$ denoting the number of relation categories. 

\subsection{Optimization Objective}
Considering that an entity pair may have multiple relations, we choose the binary cross entropy loss between the ground truth label $y_r$ and $P(r|e_1, e_2)$ as the optimization objective: 
\begin{equation}
    \begin{aligned}
    \mathcal{L} = -\sum_r \Big( y_r \cdot \log \big(P(r|e_1, e_2)\big) + & \\
    (1 - y_r) \cdot \log \big(1 - & P(r|e_1, e_2)\big) \Big). 
    \end{aligned}
\end{equation}

\begin{table*}[t]
\centering
\setlength{\tabcolsep}{15pt}
\begin{tabular}{l | c c | c c}
\toprule
\multirow{2}{*}{\textbf{Model}} & \multicolumn{2}{c|}{\textbf{Dev}} & \multicolumn{2}{c}{\textbf{Test}} \\ 
& \textbf{Ign F1} & \textbf{F1} & \textbf{Ign F1} & \textbf{F1}  \\ 
\midrule
\midrule
Bi-LSTM~\citep{docred} & 48.87 & 50.94 & 48.78 & 51.06 \\
GCNN~\citep{gcnn} & 46.22 & 51.52 & 49.59 & 51.62 \\
EoG~\citep{eog} & 45.94 & 52.15 & 49.48 & 51.82 \\
HIN~\citep{hin} & 51.06 & 52.95 & 51.15 & 53.30 \\
GCGCN~\citep{gcgcn} & 51.14 & 53.05 & 50.87 & 53.13 \\
GEDA~\citep{geda} & 51.03 & 53.60 & 51.22 & 52.97 \\
LSR~\citep{lsr} & 48.82 & 55.17 & 52.15 & 54.18 \\
GAIN~\citep{gain} & 53.05 & 55.29 & 52.66 & 55.08 \\
HeterGSAN-Recon~\citep{recon} & \underline{54.27} & \underline{56.22} & \underline{53.27} & \underline{55.23} \\
CFER (Ours) & \textbf{54.29} & \textbf{56.40} & \textbf{53.43} & \textbf{55.75} \\
\midrule
BERT-Two-Step~\citep{bert_two_step} & - & 54.42 & - & 53.92 \\
GEDA-BERT$_{\text{Base}}$~\citep{geda} & 54.52 & 56.16 & 53.71 & 55.74 \\
HIN-BERT$_{\text{Base}}$~\citep{hin} & 54.29 & 56.31 & 53.70 & 55.60 \\
GCGCN-BERT$_{\text{Base}}$~\citep{gcgcn} & 55.43 & 57.35 & 54.53 & 56.67 \\
CorefBERT$_{\text{Base}}$~\citep{coref_bert} & 55.32 & 57.51 & 54.54 & 56.96 \\
ERICA-BERT$_{\text{Base}}$~\citep{erica} & 56.7~~ & 58.8~~ & 55.9~~ & 58.2~~ \\
LSR-BERT$_{\text{Base}}$~\citep{lsr} & 52.43 & 59.00 & 56.97 & 59.05 \\
MIUK~\citep{miuk} & 58.27 & 60.11 & 58.05 & 59.99 \\
HeterGSAN-Recon-BERT$_{\text{Base}}$~\citep{recon} & 58.13 & 60.18 & 57.12 & 59.45 \\
GAIN-BERT$_{\text{Base}}$~\citep{gain} & \underline{59.14} & \underline{61.22} & \underline{59.00} & \underline{61.24} \\
CFER-BERT$_{\text{Base}}$ (Ours) & \textbf{59.23} & \textbf{61.41} & \textbf{59.16} & \textbf{61.28} \\
\midrule
ERICA-RoBERTa$_{\text{Base}}$~\citep{erica} & 56.3~~ & 58.6~~ & 56.6~~ & 59.0~~ \\
CorefRoBERTa$_{\text{Large}}$~\citep{coref_bert} & \underline{57.84} & \underline{59.93} & \underline{57.68} & \underline{59.91} \\
CFER-RoBERTa$_{\text{Large}}$ (Ours) & \textbf{60.91} & \textbf{62.80} & \textbf{60.70} & \textbf{62.76} \\
\bottomrule
\end{tabular}
\caption{Main evaluation results on DocRED. \textbf{Bold} denotes the best result. \underline{Underline} denotes the second-best result. }
\label{tab:docred_rlt}
\end{table*}

\begin{table}[t]
\centering
\setlength{\tabcolsep}{2pt}
\begin{tabular}{l | c c c}
\toprule
\textbf{Model} & \textbf{F1} & \textbf{Intra-F1} & \textbf{Inter-F1} \\
\midrule
GCNN~\citep{gcnn} & 58.6 & - & - \\
LSR~\citep{lsr} & 61.2 & 66.2 & 50.3 \\
BRAN~\citep{barn} & 62.1 & - & - \\
EoG~\citep{eog} & 63.6 & 68.2 & 50.9 \\
LSR w/o MDP Nodes & \underline{64.8} & \underline{68.9} & \underline{53.1} \\
CFER (Ours) & \textbf{65.9} & \textbf{70.7} & \textbf{57.8} \\
\bottomrule
\end{tabular}
\caption{Main evaluation results on the biomedical dataset CDR. }
\label{tab:cdr}
\end{table}

\section{Experiments and Analysis}

\subsection{Dataset}
We evaluate our model on two document-level RE datasets. 
\textbf{DocRED}~\citep{docred} is a large-scale human-annotated dataset constructed from Wikipedia and Wikidata~\citep{wikidata}, and is currently the largest human-annotated dataset for general domain document-level RE. 
It contains $5,053$ documents, $132,375$ entities and $56,354$ relational facts divided into $96$ relation categories. 
Among the annotated documents, $3,053$ are for training, $1,000$ are for development, and $1,000$ are for testing. 
Chemical-Disease Reactions~(\textbf{CDR})~\citep{cdr} is a popular dataset in the biomedical domain containing 500 training examples, 500
development examples, and 500 testing examples. 

\subsection{Experimental Settings}
We tune hyper-parameters on the development set. 
Generally, we use AdamW~\citep{adamw} as the optimizer, and use the DCGCN consisting of two blocks with 4 sub-layers in each block. 
On CDR, we use a biomedical domain pre-trained model BioBERT~\citep{biobert} as the text encoding module. 
Details of hyper-parameters such as learning rate, batch size, dropout rate, hidden dimension, and others for each version of CFER are shown in Appendix A. 

Following previous works, for DocRED, we choose micro \textbf{F1} and \textbf{Ign F1} as evaluation metrics. 
For CDR, we choose micro \textbf{F1}, \textbf{Intra-F1}, and \textbf{Inter-F1} as metrics. 
Ign F1 denotes F1 excluding relational facts that appear in both the training set and the development or testing set. 
Intra- and Inter-F1 denote F1 of relations expressed within and across sentences, respectively. 
We determine relation-specific thresholds $\delta_r$ based on the micro F1 on the development set. 
With these thresholds, we classify a triplet $(e_1, r, e_2)$ as positive if $P(r|e_1, e_2) > \delta_r$ or negative otherwise. 

\subsection{Main Results}

To obtain the main evaluation results, we run each version of CFER three times, take the median F1 on the development set to report, and test the corresponding model on the test set. 
Table~\ref{tab:docred_rlt} shows the main evaluation results on DocRED. 
We compare CFER with 21 existing models on three tracks: 
(1) \textbf{non-PTM track:}
models on this track do not use Pre-Trained Models (PTM). 
\textbf{GCNN}~\citep{gcnn}, \textbf{EoG}~\citep{eog}, \textbf{GCGCN}~\citep{gcgcn}, \textbf{GEDA}~\citep{geda}, \textbf{LSR}~\citep{lsr},  \textbf{GAIN}~\citep{gain}, and \textbf{HeterGSAN-Recon} are graph-based models that leverage GNN to encode nodes. 
\textbf{Bi-LSTM}~\citep{docred} and HIN~\citep{hin} are text-based models without constructing a graph. 
(2) \textbf{BERT track:}
models on this track are based on BERT$_\text{Base}$~\citep{bert}. 
\textbf{GEDA-BERT$_{\text{Base}}$}, \textbf{HIN-BERT$_{\text{Base}}$}, \textbf{GCGCN-BERT$_{\text{Base}}$}, \textbf{LSR-BERT$_{\text{Base}}$}, \textbf{HeterGSAN-Recon-BERT$_{\text{Base}}$}, and \textbf{GAIN-BERT$_{\text{Base}}$} are BERT-versions of corresponding non-PTM models. 
\textbf{BERT-Two-Step}~\citep{bert_two_step} and \textbf{MIUK}~\citep{miuk} are two text-based models. 
\textbf{CorefBERT$_{\text{Base}}$}~\citep{coref_bert} and \textbf{ERICA-BERT$_{\text{Base}}$}~\citep{erica} are two pre-trained models. 
(3) \textbf{RoBERTa track:}
models on this track are based on RoBERTa~\citep{roberta}: \textbf{ERICA-RoBERTa$_{\text{Base}}$}~\citep{erica} and \textbf{CorefRoBERTa$_{\text{Large}}$}~\citep{coref_bert}.  
From Table~\ref{tab:docred_rlt}, we observe that 
on all three tracks, CFER achieves the best performance and significantly outperforms most of the existing models. 
Further, we find that graph-based models generally have significant advantages over text-based models in document-level RE. 

Table~\ref{tab:cdr} shows the main evaluation results on CDR. 
Besides \textbf{GCNN}, \textbf{LSR}, \textbf{EoG} mentioned above, we compare two more models: \textbf{BRAN}~\citep{barn}, a text-based method, and \textbf{LSR w/o MDP Nodes}, a modified version of LSR. 
From Table~\ref{tab:cdr} we find that CFER outperforms all the baselines, especially on Inter-F1. 
This suggests that CFER has stronger advantages in modeling inter-sentence interactions. 
Note that the modified version of LSR performs better than full LSR. 
This may imply that LSR does not always work on all datasets, while CFER does not have this problem. 

\subsection{Model Analysis}

\noindent\textbf{Ablation study: }
To verify the effectiveness of each module in CFER, we show ablation experiment results of CFER-GloVe on the development set of DocRED in Table~\ref{tab:ablation_rlt}. 
\textbf{Firstly}, to explore the effectiveness of fine-level representations, we modify our model in three ways: 
(1) Replace the attention aggregator by a simple mean aggregator (denoted by \textit{- Attention Aggregator}). 
(2) Replace all used shortest paths by a random shortest path (denoted by \textit{- Multiple Paths}). 
(3) Remove the whole fine-level representations from the final classification (denoted by \textit{- Fine-level Repr.}). 
F1 scores of these three modified models decrease by 1.48, 2.72, and 4.12, respectively. 
This verifies that our attention aggregator for multiple paths has the ability to selectively aggregate useful information thus producing effective fine-level representations. 
\textbf{Secondly}, to explore the effectiveness of coarse-level representations, we modify our model in two ways: 
(1) Remove the whole coarse-level representations from the final classification (denoted by \textit{-  Coarse-level Repr.}). 
(2) Remove DCGCN blocks (denoted by \textit{- DCGCN Blocks}). 
F1 scores of these two modified models decrease by 1.14 and 2.08, respectively. 
This verifies that DCGCN has the ability to capture rich local and global contextual information, thus producing high-quality coarse-level representations that benefit relation extraction. 
\textbf{Finally}, we remove both coarse- and fine-level representation modules (denoted by \textit{-  Both-level Modules}). 
This operation makes our model degenerate into a simple version similar to Bi-LSTM. 
As a result, this simple version achieves a similar performance to the Bi-LSTM baseline as expected. 
This suggests that our text encoding module is not much different from the Bi-LSTM baseline, and the performance improvement is totally introduced by our coarse-to-fine strategy. 

\begin{figure}[t]
\centering
\includegraphics[width=0.99\linewidth]{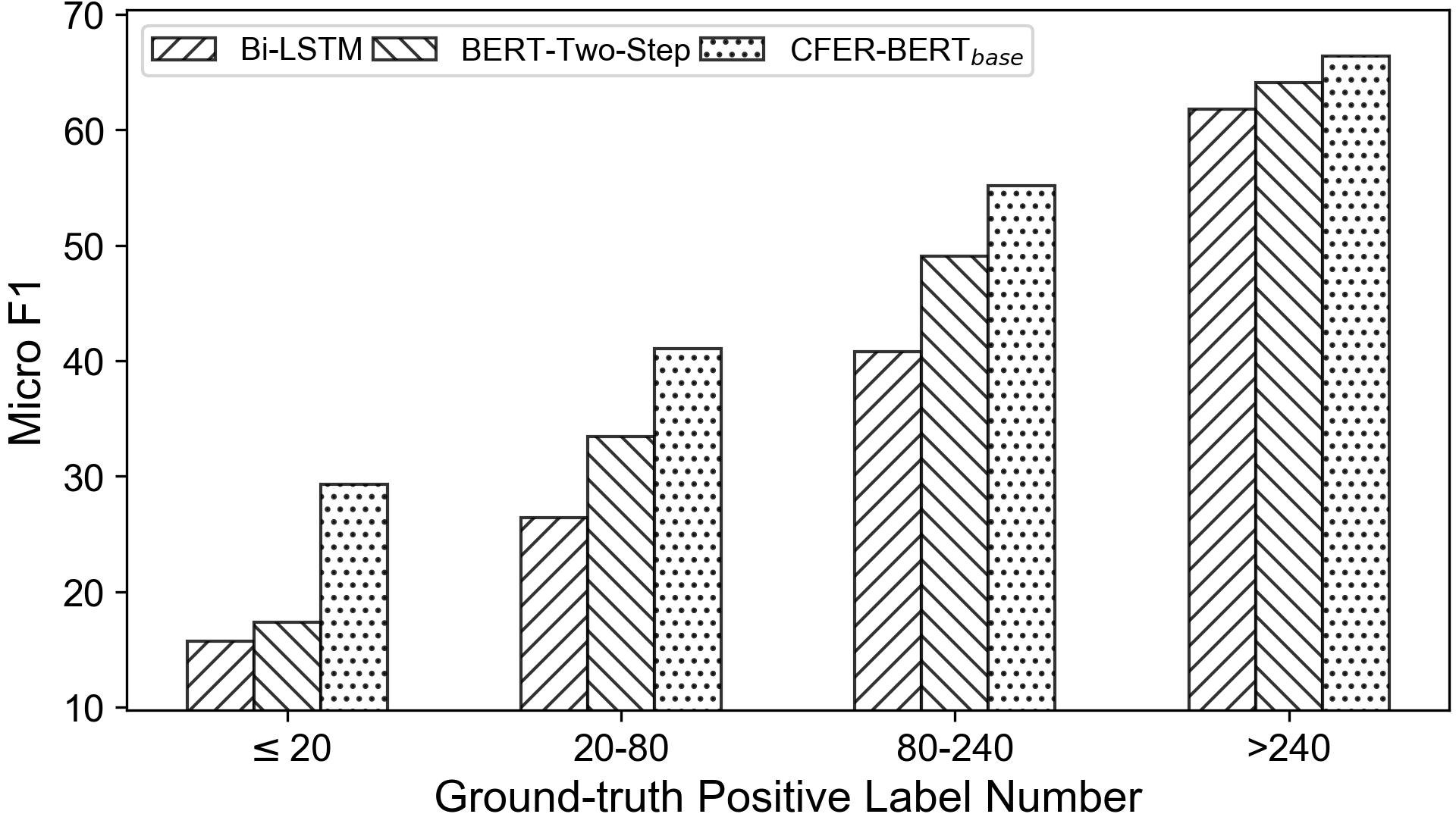}
\caption{
Micro F1 on different categories of relations. 
CFER makes more improvement on long-tail relations with fewer positive labels. 
}
\label{fig:relation_f1}
\end{figure}

\noindent\textbf{Robustness to the uneven label distribution: }
To analyze our performance on different relations, we divide 96 relations in DocRED into 4 categories according to their ground-truth positive label numbers in the development set. 
We show micro F1 on each category achieved by two baseline models and CFER in Figure~\ref{fig:relation_f1}. 
As shown in the figure, compared to Bi-LSTM, BERT-Two-Step makes more improvement on relations with more than 20 positive labels (denoted by major relations), but less improvement ($10.15\%$) on long-tail relations with less than or equal to 20 positive labels. 
By contrast, keeping the improvement on major relations, our model makes a much more significant improvement ($86.18\%$) on long-tail relations. 
With high performance on long-tail relations, our model narrows the F1 gap between major and long-tail relations from $46.05$ (Bi-LSTM) to $38.12$. 
This suggests that our model is more robust to the uneven label distribution.

\noindent\textbf{Ability to model long-distance interactions: }
To analyze our performance on entity pairs with different distances, we divide all entity pairs in CDR into 3 categories according to their sentence distance, i.e., the number of sentences that separate them. 
Specifically, the distance of an intra-sentence entity pair is 0. 
Table~\ref{tab:distance} shows the micro F1 on entity pairs with different sentence distances. 
From the table, we find that as the distance increases, the performance of CFER does not decrease much, and even increases from [4, 8) to [8, ). 
This validates again the ability of CFER to model long-distance interactions between entity pairs. 

\subsection{Case Study}

\begin{figure*}[ht]
\centering
\includegraphics[width=0.93\linewidth]{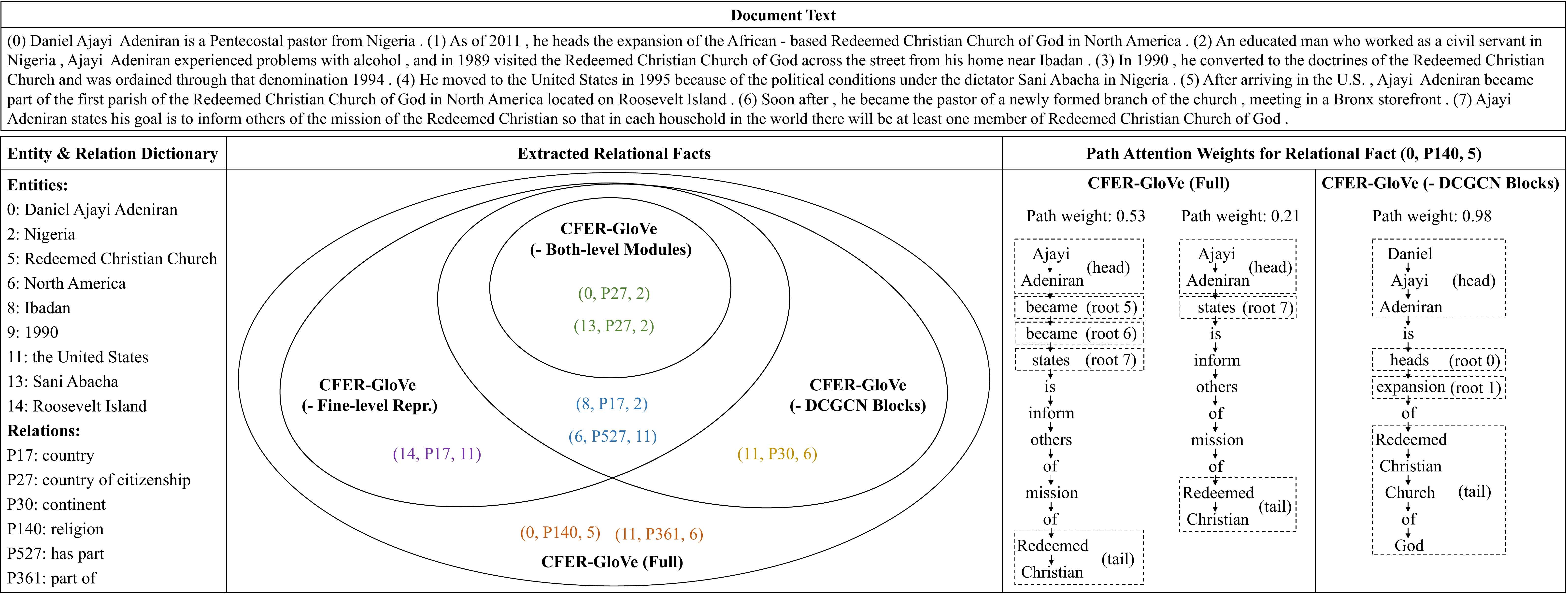}
\caption{
An extraction case from the development set. 
We show the relational facts extracted by four versions of CFER-GloVe. 
For CFER-GloVe (Full) and CFER-GloVe (- DCGCN Blocks), we additionally show the paths along with their attention weights used for producing fine-level representations of entity 0 (Daniel Ajayi Adeniran) and entity 5 (Redeemed Christian Church). 
}
\label{fig:case_study}
\end{figure*}

Figure~\ref{fig:case_study} shows an extraction case selected from the development set. 
In this case, CFER-GloVe (- Both-level Modules), a simple version similar to the Bi-LSTM baseline, extracts only two simple relational facts. 
CFER-GloVe (- Fine-level Repr.) extracts three more relational facts since it adopts DCGCN blocks to integrate richer local and global contextual information. 
CFER-GloVe (- DCGCN Blocks) makes use of path information to model long-distance entity interactions, and also extracts three more relational facts compared to CFER-GloVe (- Both-level Modules). 
CFER-GloVe (Full) combines all the advantages of its modified versions. 
As a result, it extracts the most relational facts. 

To reveal how our coarse-to-fine strategy works, we further analyze the relational fact (0, P140, 5), which is extracted by only CFER (Full). 
In Figure~\ref{fig:case_study}, for CFER (Full) and CFER (- DCGCN Blocks), we additionally show several high-weight paths along with their attention weights used for producing fine-level representations. 
From the shown case, we can find that CFER (Full) gives relatively smooth weights to its high-weight paths. 
This enables it to aggregate richer path information from several useful paths. 
In fact, CFER (Full) successfully aggregates both intra-sentence (within sentence 7) and inter-sentence (across sentences 5, 6, and 7) information. 
By contrast, without the guidance of global contextual information, CFER (- DCGCN Blocks) learns extremely unbalanced weights and pays almost all its attention to a sub-optimal path. 
As a result, it fails to extract the P140 relation. 
As for CFER-GloVe (- Fine-level Repr.) and CFER-GloVe (- Both-level Module), they do not consider any path information. 
Therefore, it is hard for them to achieve as good performance as CFER (Full). 

\begin{table}[t]
\centering
\setlength{\tabcolsep}{10pt}
\begin{tabular}{l | c}
\toprule
\textbf{Model} & \textbf{F1}\\
\midrule
\midrule
CFER-GloVe (Full) & 56.40 (-0.00) \\
\midrule
- Attention Aggregator & 54.92 (-1.48) \\
- Multiple Paths & 53.68 (-2.72) \\
- Fine-level Repr. & 52.28 (-4.12) \\
\midrule
- Coarse-level Repr. & 55.26 (-1.14) \\
- DCGCN Blocks & 54.32 (-2.08) \\
\midrule
- Both-level Modules & 51.67 (-4.73) \\
\bottomrule
\end{tabular}
\caption{Ablation experiments on the development set of DocRED. }
\label{tab:ablation_rlt}
\end{table}

\begin{table}[t]
\centering
\setlength{\tabcolsep}{10pt}
\begin{tabular}{l | c c c}
\toprule
\textbf{Distance} & [0, 4) & [4, 8) & [8, ) \\
\midrule
\textbf{Micro F1} & 66.3 & 59.8 & 61.5 \\
\bottomrule
\end{tabular}
\caption{Micro F1 on entity pairs with different sentence distances. }
\label{tab:distance}
\end{table}

\section{Related Work}

Most of existing document-level RE models are \textbf{graph-based}. 
They usually construct a document-level graph, which represents words as nodes and uses edges between them to capture document-aware interactions. 
\citet{quirk_and_poon} design feature templates to extract features from multiple paths for classification. 
\citet{gcnn} use a labeled edge GCN~\citep{labelled_edge_gcn} to integrate information. 
\citet{aggcn} and \citet{lsr} apply graph convolutional networks to a complete graph with iteratively refined edges weights. 
\citet{eog} propose an edge-oriented model that represents paths through a walk-based iterative inference mechanism. 
\citet{recon} propose a reconstructor to model path dependency between an entity pair. 
\citet{geda} propose to improve inter-sentence reasoning by characterizing interactions between sentences and relation instances. 
\citet{miuk} leverage knowledge graphs in document-level RE. 
\citet{gcgcn} propose a context-enhanced model to capture global context information. 
\citet{gain} design a double graph to cope with document-level RE. 

Besides graph-based methods, there are also \textbf{text-based} methods without constructing a graph. 
\citet{barn} adopt a transformer~\citep{transformer} to encode the document. 
\citet{bert_two_step} adopt BERT~\citep{bert} to encode the document and predict the existence of relations before predicting the specific relations. 
\citet{hin} design a hierarchical architecture to make full use of information from several levels. 
\citet{coref_bert} attempt to explicitly capture relations between coreferential noun phrases to coherently comprehend the whole document.
\citet{erica} use contrastive learning to obtain a deeper understanding of entities and relations in text. 

\section{Conclusion}

In this paper, we propose CFER with a coarse-to-fine strategy to learn comprehensive representations for document-level RE. 
Our model integrates global contextual information and models long-distance interactions between the target entity pair simultaneously, thus addressing the disadvantages that existing graph-based models suffer from. 
Experimental results on two document-level RE datasets, DocRED and CDR, show that CFER outperforms existing models. 
Further, elaborate analysis verifies the effectiveness of our coarse-to-fine strategy, and highlights the robustness of CFER to the uneven label distribution and the ability of CFER to model long-distance interactions. 
Note that our coarse-to-fine strategy is not limited to only the task of document-level RE. 
It has the potential to be applied to a variety of other NLP tasks.

\small
\bibliographystyle{named} 
\bibliography{ijcai21}

\appendix
\appendixpage

\section{Details of Hyper-parameters}

We search the best hyper-parameters based on F1 on the development set. 
Generally, for all of CFER-GloVe, CFER-BERT$_{\text{Base}}$, CFER-RoBERTa$_{\text{Large}}$ and CFER for CDR, we use AdamW with $\beta1=0.9$, $\beta2=0.999$, $\epsilon=1e-6$, $\text{weight decay}=0.0001$ as the optimizer, apply exponential moving average on all parameters with a decay rate of $0.9999$, use ReLU as the activation function, and use the DCGCN consisting of two blocks with 4 sub-layers in each block. 
We adopt a slanted triangular scheduling strategy for learning rate, which first linearly increases the learning rate from $0$ to the peak value in the first $10\%$ steps (warm-up steps), and then linearly decreases it to $0$ in remaining steps. 
For other key hyper-parameters, we state the values tried and the finally selected value for four models separately as follows. 

For CFER-GloVe: 
(1) We search the peak learning rate for all modules in $\{1e-3, 1e-4\}$, and finally choose $1e-3$. 
(2) We search the batch size in $\{8, 16, 32\}$, and finally select $16$.
(3) We search the dropout rate for DCGCN modules in $\{0.2, 0.4, 0.6\}$, and finally select $0.4$.
(4) We search the dropout rate for other modules in $\{0.2, 0.4, 0.6\}$, and finally select $0.2$.
(5) We set the embedding dimension to $300$, the same as the dimension of used GloVe embeddings. 
(6) We search the hidden size in $\{100, 300, 512\}$, and finally select $300$. 
(7) For each hyper-parameter configuration, we train $300$ epochs and select the best F1 achieved during these $300$ epochs to evaluate the performance under this configuration. 

For CFER-BERT$_{\text{Base}}$: 
(1) We search the peak learning rate for BERT modules in $\{1e-4, 5e-5, 1e-5\}$, and finally select $1e-5$.
(2) We search the peak learning rate for the other modules in $\{1e-3, 5e-4, 1e-4\}$, and finally select $1e-3$. 
(3) We search the batch size in $\{8, 16, 32\}$, and finally select $32$.
(4) We search the dropout rate for DCGCN modules in $\{0.2, 0.4, 0.6\}$, and finally select $0.6$.
(5) We search the dropout rate for other modules in $\{0.2, 0.4, 0.6\}$, and finally select $0.2$. 
(6) We search the hidden size in $\{300, 512, 768\}$, and finally select $512$. 
(7) For each hyper-parameter configuration, we train $300$ epochs and select the best F1 achieved during these $300$ epochs to evaluate the performance under this configuration. 

For CFER-RoBERTa$_{\text{Large}}$: 
(1) We search the peak learning rate for RoBERTa modules in $\{1e-4, 5e-5, 1e-5\}$, and finally select $1e-5$. 
(2) We search the peak learning rate for the other modules in $\{1e-3, 5e-4, 1e-4\}$, and finally select $1e-3$. 
(3) We search the batch size in $\{8, 16, 32\}$, and finally select $32$.
(4) We search the dropout rate for DCGCN modules in $\{0.2, 0.4, 0.6\}$, and finally select $0.6$.
(5) We search the dropout rate for other modules in $\{0.2, 0.4, 0.6\}$, and finally select $0.2$.
(6) We search the hidden size in $\{512, 768, 1024\}$, and finally select $1024$. 
(7) For each hyper-parameter configuration, we train $300$ epochs and select the best F1 achieved during these $300$ epochs to evaluate the performance under this configuration. 

For CFER for CDR: 
(1) We search the peak learning rate for BioBERT modules in $\{1e-4, 5e-5, 1e-5\}$, and finally select $1e-5$. 
(2) We search the peak learning rate for the other modules in $\{1e-3, 5e-4, 1e-4\}$, and finally select $1e-4$. 
(3) We search the batch size in $\{4, 8, 16\}$, and finally select $4$.
(4) We search the dropout rate for DCGCN modules in $\{0.2, 0.4, 0.6\}$, and finally select $0.6$.
(5) We search the dropout rate for other modules in $\{0.2, 0.4, 0.6\}$, and finally select $0.2$.
(6) We search the hidden size in $\{512, 768, 1024\}$, and finally select $1024$. 
(7) For each hyper-parameter configuration, we train $100$ epochs and select the best F1 achieved during these $100$ epochs to evaluate the performance under this configuration. 

\end{document}